\documentclass[runningheads]{llncs}
\usepackage{cite}
\usepackage{amsmath,amssymb,amsfonts}
\usepackage{graphicx}
\usepackage{widetable}
\usepackage{multirow}
\usepackage{booktabs}

\begin{document}
\title{Automated Classification of General Movements in Infants Using Two-stream Spatiotemporal Fusion Network}
\titlerunning{Automated Classification of General Movements in Infants}
\author{Yuki Hashimoto\inst{1}\thanks{indicates equal contribution} \and
Akira Furui\inst{1}\protect\footnotemark[1] \and
Koji Shimatani\inst{2} \and
Maura Casadio\inst{3} \and
Paolo Moretti\inst{4} \and
Pietro Morasso\inst{5} \and
Toshio Tsuji\inst{1}}

\authorrunning{Hashimoto et al.}

\institute{Hiroshima University, Hiroshima, Japan\\
\email{\{yukihashimato,akirafurui,toshiotsuji\}@hiroshima-u.ac.jp} \and
Prefectural University of Hiroshima, Hiroshima, Japan \and
University of Genoa, Genova, Italy \and
Istituto G.Gaslini Pediatric Hospital, Genova, Italy \and
Italian Institute of Technology, Genova, Italy}
\maketitle
\begin{abstract}
The assessment of general movements (GMs) in infants is a useful tool in the early diagnosis of neurodevelopmental disorders. However, its evaluation in clinical practice relies on visual inspection by experts, and an automated solution is eagerly awaited. Recently, video-based GMs classification has attracted attention, but this approach would be strongly affected by irrelevant information, such as background clutter in the video. Furthermore, for reliability, it is necessary to properly extract the spatiotemporal features of infants during GMs. In this study, we propose an automated GMs classification method, which consists of preprocessing networks that remove unnecessary background information from GMs videos and adjust the infant's body position, and a subsequent motion classification network based on a two-stream structure. The proposed method can efficiently extract the essential spatiotemporal features for GMs classification while preventing overfitting to irrelevant information for different recording environments. We validated the proposed method using videos obtained from 100 infants. The experimental results demonstrate that the proposed method outperforms several baseline models and the existing methods.

\end{abstract}

\section{Introduction}
Neurodevelopmental disorders (NDs) are impairments of brain function and nervous system development.
Patients with NDs may experience real-life difficulties due to biases and problems in cognition, movement, social skills, and attention.
Meanwhile, the brain is capable of modifying the structure and function of the central nervous system, known as plasticity, and the younger the brain, the greater the plasticity.
Therefore, early detection of the signs of NDs followed by effective intervention is crucial.

General movements (GMs) assessment is one of the most predictive and valid methods for early detection of NDs~\cite{prechtl1990qualitative,einspieler2005prechtl}.
GMs are spontaneous movements observed during early infancy that reflect the state of the central nervous system.
In general, experts observe video recordings of infants and assess the quality of GMs, especially abnormalities in movements or the lack of specific movement patterns.
Qualitative abnormalities in GMs are closely related to the risk of cerebral palsy and various NDs~\cite{ferrari1990qualitative,beccaria2012poor}.
However, GMs assessment relies on visual inspection, which requires a high level of expertise and places a heavy burden on experts because of the need for prolonged observation; thus, an automated solution for GMs classification is eagerly awaited.

Recently, automated classification of GMs based on video recognition has attracted significant attention~\cite{adde2010early,adde2018characteristics,schmidt2019general,tsuji2020markerless,mccay2020abnormal,nguyen2021spatio}.
In this approach, the following is important.
(i) Separation of infant and irrelevant information: The cost of GMs recording in a medical setting is very high, and it is not easy to obtain a large dataset.
Additionally, the number of high-risk infants with abnormal GMs is generally small and often concentrated in specific medical institutions, which can lead to bias in the recording environment depending on the types of GMs.
The use of these recordings possesses the risk of overfitting the recognition model to irrelevant information, such as background clutter (e.g., wrinkles in the sheets or bed frame) and differences in the relative body sizes in the videos.
(ii) Appropriate extraction of spatiotemporal features: Experts assess GMs by comprehensively evaluating the infant's spatial features (appearance) and their temporal evolution (movement).
Temporal features such as motor intensity and velocity would be the most important factors characterizing GMs, and abnormal GMs lack fluency and complexity~\cite{Einspieler2016-nq}.
Spatial features such as the body shape and posture may also contribute to the reliability of GMs assessment~\cite{hesse2018learning}.
Therefore, if we can extract and integrate the effective spatial and temporal components from a video while removing irrelevant visual artifacts, an accurate and reliable recognition architecture for GMs classification can be developed.

In this study, we propose an automated GMs classification method that can remove irrelevant information from videos and efficiently learn the spatiotemporal features of infants during spontaneous movements.
The proposed method consists of preprocessing networks and a motion classification network.
First, the preprocessing networks receive the measured video and extract only the infant's body area from each frame using the mask obtained by salient object detection.
The relative size and angle of the infant's body in the video are then unified among different individuals utilizing the joint coordinates obtained from the pose estimation model.
Subsequently, we construct a single-frame image and stacked multi-frame optical flow from the preprocessed video, and the motion classification network with a two-stream architecture extracts temporal and spatial features to predict the types of GMs.

The main contributions of this study are as follows:
\begin{itemize}
\item We introduce preprocessing networks to remove irrelevant information from the videos automatically, thereby preventing overfitting to non-essential elements for GMs classification.
\item We introduce a motion classification network based on a two-stream architecture. By fusing a spatial stream with a single-frame image and a temporal stream with a multi-frame optical flow, this network can efficiently learn the spatiotemporal features that characterize the GMs.
\end{itemize}

\section{Related work}
GMs are whole-body spontaneous movements that appear 8--9 weeks after fertilization and are observed until 15--20 weeks of corrected age, giving an impression of fluency and grace~\cite{prechtl1990qualitative,einspieler2005prechtl}.
These movements are generated by a central pattern generator that is believed to reside between the brainstem and spinal cord and reflect the state of the infant's nervous system.
Writhing movements (WMs) observed in the post-term period are characterized by elliptical movements of the limbs, sometimes with large extensions of the upper limbs~\cite{einspieler2005prechtl}.
At the post-term ages of 6--9 weeks, WMs gradually disappear and fidgety movements (FMs) emerge, in which the head, trunk, and limbs move in all directions in tiny movements~\cite{prechtl1997state}.
In contrast, movements that are absent or qualitatively different from normal movements are considered abnormal GMs.
For example, poor repertoire GMs (PR) are classified if the sequence of successive movement components is monotonous, and the intensity, velocity, and range of motion lack the normal variability as seen in WMs~\cite{Einspieler2016-nq}.
Qualitative abnormalities of these GMs are a good predictor of cerebral palsy, autism spectrum disorders, and delayed cognitive and language development~\cite{beccaria2012poor,Einspieler2016-nq}.

Video-based GMs classification methods that do not interfere with infants' natural movements have been studied to develop automated early screening tools for NDs.
Various methods have been developed to design features of spontaneous movements in infants based on image processing~\cite{maggi2017new,tacchino2021spontaneous}, and some of these methods have been applied to GMs classification~\cite{adde2010early,adde2018characteristics,tsuji2020markerless}.
However, these feature engineering-based approaches have difficulty in comprehensively capturing the discriminative features specific to GMs classification.
Recently, the automatic extraction of spatiotemporal features that characterize GMs based on deep neural networks, including a convolutional neural network (CNN)-based method~\cite{schmidt2019general} and a combination of pose estimation and attention mechanisms~\cite{nguyen2021spatio}, has been shown to be effective.
These video-based methods, however, involve the problem of being strongly affected by irrelevant information, such as background clutter and relative positions of the infant's body.
The pose-based approach~\cite{chambers2020computer,nguyen2021spatio} may mitigate this problem, but its performance depends on the accuracy of the pose estimation algorithm.
Therefore, we propose a video-based GMs classification method that can remove irrelevant, distracting information from the video and learn spatial and temporal features based on a two-stream architecture to extract effective features related to GMs.

\section{Proposed GMs classification method}
\begin{figure}[t]
\centering
\includegraphics[width=\linewidth]{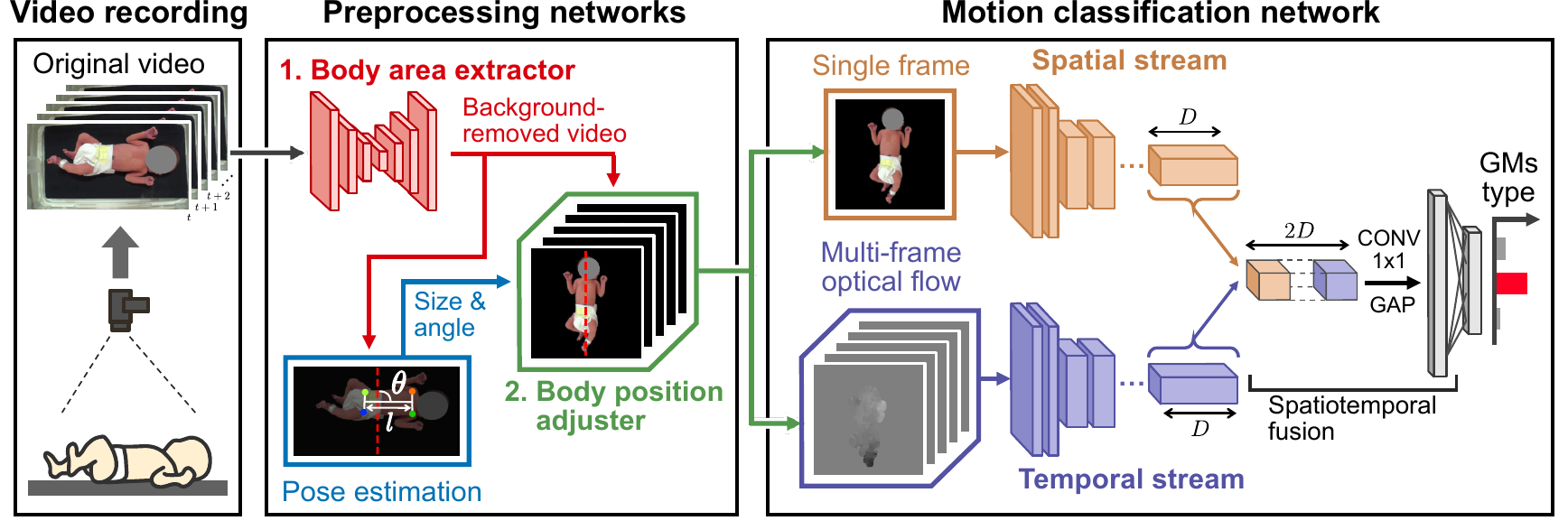}
\caption{Overview of the proposed GMs classification method. The abbreviations GAP and CONV $1 \times 1$ denote global average pooling and pointwise convolution, respectively.}
\label{overview}
\end{figure}

Fig.~\ref{overview} shows an overview of the proposed method.
The proposed method consists of the preprocessing networks and motion classification network.
The preprocessing networks extract the infant's body area and adjust the relative body size and position in the image to remove the effects of visual artifacts.
The motion classification network predicts the types of GMs based on a two-stream architecture to efficiently extract the spatial and temporal features of infants during GMs.

\subsection{Preprocessing networks}
In general, video recording of GMs is performed using a single RGB camera fixed above the bed to capture the movements of an infant lying supine, which has two difficulties.
First, the video contains irrelevant information other than the infant's body, such as wrinkles in sheets, the bed frame, and the floor.
Second, the relative scale and orientation of the infant with respect to the camera may differ for each recording period.
It is difficult to completely standardize these conditions among different medical institutions.
In the preprocessing networks, the former problem is solved by a body area extractor, and the latter problem is solved by a body position adjuster.

\textbf{Body area extractor} separates the infant's body and background information.
We use U$^2$-Net~\cite{qin2020u2} pretrained on the DUST-TR dataset, which has shown good segmentation results for various tasks.
We apply the segmentation based on U$^2$-Net for each frame and extract the infant's body by masking the output saliency maps to the original video frames.
\textbf{Body position adjuster} unifies the scale and orientation of the infant across different videos.
This adjuster utilizes OpenPose~\cite{cao2017realtime}, a pose estimation network.
Out of the 18 joints output from this network, we use four coordinates for both shoulders and hips.
The vector passing through the midpoints of both shoulders and hips is defined as the approximate body-axis direction and is used for adjustment.

The adjustment is performed in the following two steps.
In the first step, pose estimation is applied to the first frame of the video and obtain the angle of the body axis relative to the vertical direction of the video.
We then rotate all the video frames using this angle to roughly align the body orientation.
In the second step, pose estimation is applied again to each frame of the rotated video, and the following three quantities at frame $t$ are calculated: the angle $\theta_t$ of the body axis, center coordinate $\mathbf{c}_t$ of the line segment connecting the midpoints of both shoulders and hips, and length $l_t$ of the line segment.
To consistently adjust the video frames, the averages of the quartile range of the above quantities are calculated for every frame and defined as $\bar{\theta}$, $\bar{\mathbf{c}}$, and $\bar{l}$, respectively.
The body orientation is readjusted by rotating all the frames using $\bar{\theta}$, and each video frame is cropped or uncropped with a square such that $\bar{\mathbf{c}}$ coincides with the center position of the video frame after rotation. 
We set the length of one side of the square to $R = 3 \bar{l} / \alpha$, where $\alpha > 0$ is an arbitrary parameter that fits the infant's body in the video at a reasonable scale.
Finally, all video frames are resized by $W \times H$ because $R$ has a different value for each video.

\subsection{Motion classification network with two-stream architecture}
We construct the motion classification network based on the two-stream approach~\cite{simonyan2014two,feichtenhofer2016convolutional} consisting of two CNNs: spatial and temporal streams.
The feature maps in the output layer of each stream are fused, and the softmax scores are finally calculated through the fully connected layer.

The spatial and temporal streams receive a single-frame image and a stacked multi-frame optical flow, respectively.
Here, the optical flow is a set of displacement vector fields between consecutive pairs of frames, $t$ and $t+1$, reflecting the infant's motion information between frames.
Dense optical flow is calculated from each frame of the RGB video $\mathbf{x}_t \in \mathbb{R}^{W \times H \times 3}$ output from the preprocessing networks using the Farneback method~\cite{farneback2003two}, and the horizontal and vertical components of the vector field, $\mathbf{d}^\mathrm{h}_t\in\mathbb{R}^{W\times H}$ and $\mathbf{d}^\mathrm{v}_t\in\mathbb{R}^{W\times H}$, are extracted. 

To represent motion across a sequence of frames, the input of each stream is given by a temporal chunk consisting of $L$ consecutive frames. 
Each chunk is separated by $\tau$ frames. 
For the $n$-th chunk, the input of the spatial stream, ${\mathbf{x}^\mathrm{s}_n}$, is a single-frame image at the center of the chunk, as follows: 
\begin{equation}
\mathbf{x}^\mathrm{s}_n = \mathbf{x}_{(n-1)\tau + \frac{L}{2}-1}.
\end{equation}
The input of the temporal stream, $\mathbf{x}^\mathrm{t}_n \in \mathbb{R}^{W \times H \times 2L}$, is constructed by stacking the flow components $\mathbf{d}^\mathrm{h}_t$ and $\mathbf{d}^\mathrm{v}_t$ in the chunk along the channel direction:
\begin{equation}
\mathbf{x}^\mathrm{t}_{n}(2k-1) = \mathbf{d}^\mathrm{h}_{(n-1)\tau + k - 1}, \quad
\mathbf{x}^\mathrm{t}_{n}(2k) = \mathbf{d}^\mathrm{v}_{(n-1)\tau + k - 1},
\end{equation}
where $k=1,2,\ldots,L$ and $\mathbf{x}^\mathrm{t}_{n}(c)$ is the explicit element of $\mathbf{x}^\mathrm{t}_n$ in channel $c$.

For both streams, we use CNNs pretrained on a large-scale dataset.
Each CNN stream is combined after the activation function of the final convolutional layer.
Because the input of the temporal stream has a channel size of $2L$, the first layer of the temporal CNN is modified to fit the dimensionality.

The feature maps output from the spatial and temporal streams, $\mathbf{y}_n^\mathrm{s}\in\mathbb{R}^{W'\times H' \times D}$ and $\mathbf{y}_n^\mathrm{t}\in\mathbb{R}^{W'\times H' \times D}$, are combined to create the final feature vector, where $W'$, $H'$, and $D$ are the width, height, and the number of channels of the respective feature maps.
First, we stack the two feature maps at the same spatial locations across the feature channels to form $\mathbf{y}_n^\mathrm{cat}\in\mathbb{R}^{W'\times H' \times 2D}$.
We then transform $\mathbf{y}_n^\mathrm{cat}$ by pointwise convolution using the filters $\mathbf{f}\in\mathbb{R}^{1 \times 1 \times 2D \times D}$ and biases $\mathbf{b} \in \mathbb{R}^D$ to learn the correspondence between the two feature maps:
\begin{equation}
\widetilde{\mathbf{y}}_n=\mathbf{y}_n^\mathrm{cat}*\mathbf{f} + \mathbf{b},
\end{equation}
where $*$ denotes the operator for convolution.
After global average pooling is applied to the fused feature map $\widetilde{\mathbf{y}}_n$, the prediction score is calculated via a fully connected layer and softmax activation.
Because the prediction score is calculated for each temporal chunk, the video-wise prediction of the GMs class is finally determined by averaging the scores over all the chunks in the video.

\section{Experiments}    
To evaluate the validity of the proposed method, we conducted GMs classification experiments. 
In the experiments, we used a dataset of videos of infants captured in medical institutions at Japan and Italy. 
The aim of the study was fully explained to each infant’s parents, and informed consent was obtained before participation in the experiment. 
All experiments were approved by the Ethics Committee of the Gaslini Pediatric Hospital and Hiroshima University (registration numbers: IGGPM01-2013 and E-1150-2).
Our code is available at \url{https://github.com/uoNuM/two-stream-gma}.

\subsubsection{Dataset and implementation details:}
The dataset consisted of videos obtained from 100 infants; each video was captured at a frame rate of 30~fps and resolution of $1280 \times 720$~pixels.
The length of the original videos ranged from 60 to 210~s, and we clipped consecutive 60 s frames during GMs from each original video, avoiding periods of sleep and crying.
The infants' gestational ages ranged from 210 to 295 days, and their birth weight range was from 1400 to 3985 g.
GMs labels were attached to each video based on annotations by well-trained experts with GMs evaluation licenses. The resulting labels are as follows: WMs, 37; FMs, 36; and PR, 27. 
Here, WMs and FMs are normal GMs, and PR is abnormal.
To evaluate the classification performance, we performed a stratified infant-wise 5-fold cross-validation on the dataset.

We used ResNet-50~\cite{he2016deep} that was pretrained on the ImageNet dataset for each CNN stream.
Accordingly, the resize in the body position adjuster was set to $W = H = 224$, and the body scale parameter was set to $\alpha = 0.8$.
Each video was downsampled to 6 fps and the length and interval of the temporal chunks were set to $L=\tau = 30$.
The entire motion classification network was fine-tuned on our GMs dataset using the AdamW~\cite{loshchilov2017decoupled} optimizer, with a learning rate of $10^{-5}$.
During fine-tuning, we performed horizontal flipping data augmentation, which involved the inversion of the horizontal component of the optical flow.

\subsubsection{Experimental conditions:}
We compared the proposed method with two existing methods for GMs classification.
One is an image processing-based system proposed by Tsuji \textit{et al.} that uses 25 domain-dependent features calculated from background subtractions and inter-frame differences~\cite{tsuji2020markerless}.
The other is STAM~\cite{nguyen2021spatio}, which is a state-of-the-art method for infant movement classification using graph neural networks with features obtained from pose estimation as input.
Both existing methods were retrained from scratch using our GMs dataset.
As baselines, two types of action recognition networks, CNN + LSTM~\cite{donahue2015long} and C3D~\cite{tran2015learning}, were used instead of the motion classification network of the proposed method.
The preprocessing networks were also applied to these baselines.
We used ResNet-50 pretrained on ImageNet for the former CNN and the latter was pretrained on the Sports-M1 dataset.
Both were finetuned in the same way as the proposed method.
In addition, we conducted an ablation study to evaluate the effectiveness of the body area extractor and the body position adjuster in the preprocessing networks.

We used accuracy, the Matthews correlation coefficient (MCC), precision, and recall as performance measures.
For robust results, we repeated the analysis by changing the random seed five times and calculated the mean and standard deviation of each measure.
We did not perform statistical tests due to the small sample size.

\begin{table}[t]
\caption{Quantitative evaluation results (\textit{mean} $\pm$ \textit{standard deviation})
}
\label{tab:comparison}
\begin{widetable}{\textwidth}{l|cccc}
\toprule
Method & Accuracy   & MCC & Precision & Recall \\ 
\midrule
\textbf{Existing method} &\\
\:\, Tsuji \textit{et al.} \cite{tsuji2020markerless} & $0.556\pm{0.010}$ & $0.331\pm{0.016}$ & $0.556\pm{0.005}$ & $0.539\pm{0.013}$\\
\:\, STAM \cite{nguyen2021spatio} & $0.640\pm{0.032}$ & $0.408\pm{0.044}$ & $0.646\pm{0.046}$ & $0.427\pm{0.034}$ \\
\cmidrule(){1-5}
\textbf{Baseline} &\\
\:\, CNN + LSTM \cite{donahue2015long} & $0.694\pm{0.031}$ & $0.563\pm{0.051}$ & $0.621\pm{0.057}$ & $0.659\pm{0.035}$ \\
\:\, C3D \cite{tran2015learning} & $0.700\pm{0.009}$ & $0.556\pm{0.016}$ & $0.684\pm{0.012}$ & $0.678\pm{0.010}$ \\
\cmidrule(){1-5}
\textbf{Ours} &\\
\:\, Only spatial & $0.742\pm{0.029}$ & $0.628\pm{0.039}$ & $0.750\pm{0.022}$ & $0.723\pm{0.026}$ \\
\:\, Only temporal & $0.696\pm{0.019}$ & $0.551\pm{0.033}$ & $0.693\pm{0.030}$ & $0.670\pm{0.018}$ \\
\:\, \underline{Two-stream fusion} & \scalebox{0.94}[1.0]{$\mathbf{0.752\pm{0.013}}$} & \scalebox{0.94}[1.0]{$\mathbf{0.647\pm{0.015}}$} & \scalebox{0.94}[1.0]{$\mathbf{0.780\pm{0.010}}$} & \scalebox{0.94}[1.0]{$\mathbf{0.737\pm{0.011}}$} \\
\bottomrule
\end{widetable}
\end{table}

\subsubsection{Results:}
Table~\ref{tab:comparison} shows the quantitative evaluation results of each method.
Our method also shows the results of using each stream individually.
The proposed method with two-stream fusion achieves the best performance for all performance measures.
The performance decreases when the structure of the motion classification network was changed to a single stream or a baseline.
Therefore, the network based on the two-stream architecture was effective for extracting the spatiotemporal features of infants during GMs.

\begin{figure}[t]
\centering
\includegraphics[width=\linewidth]{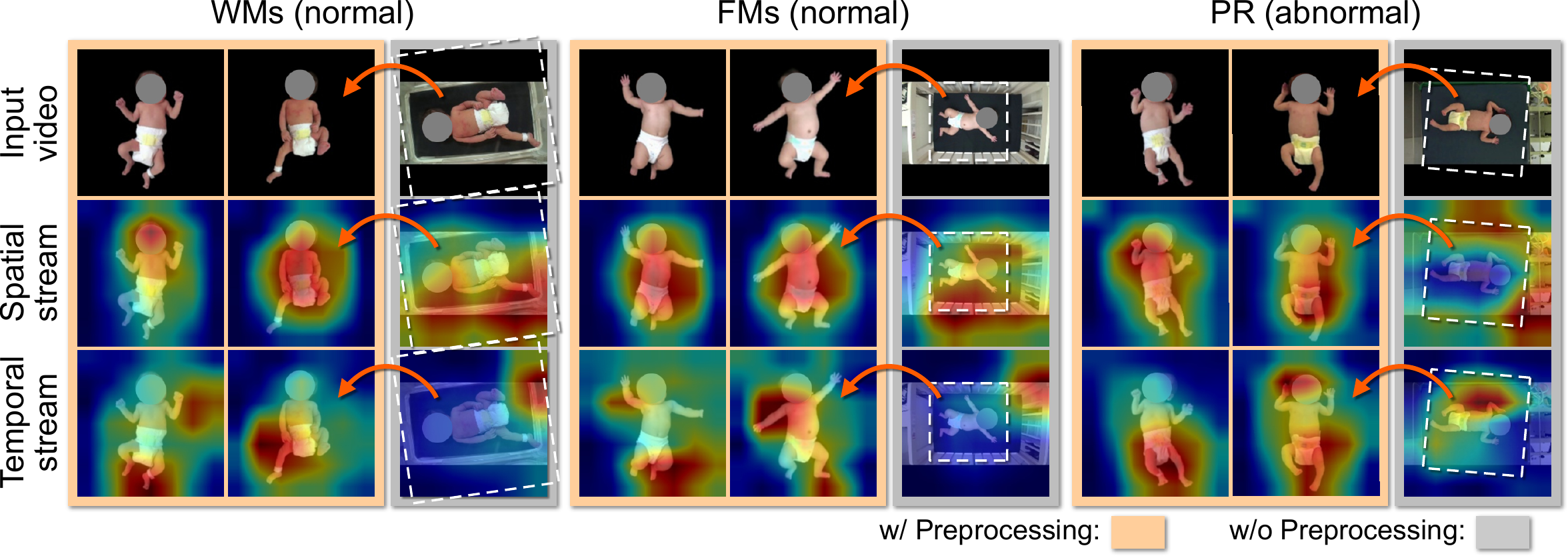}
\caption{Visualization of activation maps using Grad-CAM++. The results of the proposed method with the preprocessing networks are shown with two examples for each GMs class. The rightmost column for each class is an example without the preprocessing networks, corresponding to the second example with the preprocessing networks.}
\label{gradcam}
\end{figure}

Table~\ref{tab:ablation} shows the results of the ablation study.
The combination of each element of the preprocessing networks shows the best performance except for accuracy.
Although the performance without the preprocessing networks is also relatively high, this is because the model focuses on irrelevant information other than the infant (as shown later in Fig.~\ref{gradcam}).
Overfitting to such non-essential but class-related components may cause ostensibly high validation performance in the limited dataset.
In fact, when only the extractor is applied, such components are not referred, resulting in a loss of performance.
In contrast, incorporating the adjuster to unify the scale and orientation of the infant's body enables appropriate feature extraction and improves overall performance.

\begin{table}[t]
\caption{Results of the ablation study for the preprocessing networks with the body area extractor and body position adjuster (\textit{mean} $\pm$ \textit{standard deviation}). The absence of \checkmark\ means that the corresponding element is removed from the proposed method.}
\label{tab:ablation}
\begin{widetable}{\textwidth}{cc|cccc}
\toprule
Extractor & Adjuster & Accuracy   & MCC & Precision & Recall \\ 
\midrule
 & & \scalebox{0.94}[1.0]{$\mathbf{0.754\pm{0.016}}$} & $0.636\pm{0.024}$ & $0.760\pm{0.016}$ & $0.734\pm{0.017}$\\
\checkmark & & $0.734\pm{0.010}$ & $0.623\pm{0.013}$ & $0.742\pm{0.025}$ & $0.724\pm{0.012}$ \\
\checkmark & \checkmark & $0.752\pm{0.013}$ & \scalebox{0.94}[1.0]{$\mathbf{0.647\pm{0.015}}$} & \scalebox{0.94}[1.0]{$\mathbf{0.780\pm{0.010}}$} & \scalebox{0.94}[1.0]{$\mathbf{0.737\pm{0.011}}$} \\
\bottomrule
\end{widetable}
\end{table}

Fig.~\ref{gradcam} shows some typical examples of class activation maps using Grad-CAM++~\cite{chattopadhay2018grad}.
These activation maps were calculated for each stream in the proposed two-stream architecture.
The results demonstrate that the proposed method with the preprocessing networks adequately captures different aspects of infants during GMs in each stream.
The spatial stream focuses on the infant's overall appearance, including body shape and posture, while the temporal stream pays more attention to the limbs, which greatly influences the impression of movement. 
In contrast, the proposed method without the preprocessing networks focuses on the bed frame or floor texture, irrelevant to the infant.

\section{Conclusion}
This study proposed a classification method for GMs in infants based on spatiotemporal fusion.
In the proposed method, we introduce preprocessing networks to remove irrelevant information from the video and adjust the infant's body position.
To capture spatiotemporal feature representations of infants during GMs, we also introduced a two-stream network with a single-frame image and stacked optical flow as the input of each stream to classify the types of GMs.
The experimental results show that the proposed method outperforms existing GMs classification methods and suppresses overfitting to irrelevant elements in the video owing to the preprocessing networks.

One of the limitations of this study is that no quantitative evaluation of the reliability of the calculated optical flows has been made, which should be demonstrated in future work.
We will also introduce a disentangle representation learning to explore the spatial and temporal features that are most strongly associated with GMs assessment. 
Furthermore, we plan to extend the proposed framework to a wider range of spontaneous movement assessments, such as longitudinal assessments.

\bibliography{paper1919}

\begin{thebibliography}{10}
\providecommand{\url}[1]{\texttt{#1}}
\providecommand{\urlprefix}{URL }
\providecommand{\doi}[1]{https://doi.org/#1}

\bibitem{adde2010early}
Adde, L., Helbostad, J.L., Jensenius, A.R., Taraldsen, G., Grunewaldt, K.H.,
  St{\o}en, R.: Early prediction of cerebral palsy by computer-based video
  analysis of general movements: a feasibility study. Developmental Medicine \&
  Child Neurology  \textbf{52}(8),  773--778 (2010)

\bibitem{adde2018characteristics}
Adde, L., Yang, H., S{\ae}ther, R., Jensenius, A.R., Ihlen, E., Cao, J.y.,
  St{\o}en, R.: Characteristics of general movements in preterm infants
  assessed by computer-based video analysis. Physiotherapy Theory and Practice
  \textbf{34}(4),  286--292 (2018)

\bibitem{beccaria2012poor}
Beccaria, E., Martino, M., Briatore, E., Podest{\`a}, B., Pomero, G., Micciolo,
  R., Espa, G., Calzolari, S.: Poor repertoire general movements predict some
  aspects of development outcome at 2 years in very preterm infants. Early
  Human Development  \textbf{88}(6),  393--396 (2012)

\bibitem{cao2017realtime}
Cao, Z., Simon, T., Wei, S.E., Sheikh, Y.: Realtime multi-person {2D} pose
  estimation using part affinity fields. In: Proceedings of the IEEE Conference
  on Computer Vision and Pattern Recognition (CVPR). pp. 7291--7299 (2017)

\bibitem{chambers2020computer}
Chambers, C., Seethapathi, N., Saluja, R., Loeb, H., Pierce, S.R., Bogen, D.K.,
  Prosser, L., Johnson, M.J., Kording, K.P.: Computer vision to automatically
  assess infant neuromotor risk. IEEE Transactions on Neural Systems and
  Rehabilitation Engineering  \textbf{28}(11),  2431--2442 (2020)

\bibitem{chattopadhay2018grad}
Chattopadhay, A., Sarkar, A., Howlader, P., Balasubramanian, V.N.:
  {Grad-CAM++}: Generalized gradient-based visual explanations for deep
  convolutional networks. In: Proceedings of the IEEE Winter Conference on
  Applications of Computer Vision (WACV). pp. 839--847 (2018)

\bibitem{donahue2015long}
Donahue, J., Anne~Hendricks, L., Guadarrama, S., Rohrbach, M., Venugopalan, S.,
  Saenko, K., Darrell, T.: Long-term recurrent convolutional networks for
  visual recognition and description. In: Proceedings of the IEEE Conference on
  Computer Vision and Pattern Recognition (CVPR). pp. 2625--2634 (2015)

\bibitem{Einspieler2016-nq}
Einspieler, C., Bos, A.F., Libertus, M.E., Marschik, P.B.: The general movement
  assessment helps us to identify preterm infants at risk for cognitive
  dysfunction. Frontiers in Psychology  \textbf{7}, ~406 (2016)

\bibitem{einspieler2005prechtl}
Einspieler, C., Prechtl, H.F.: Prechtl's assessment of general movements: a
  diagnostic tool for the functional assessment of the young nervous system.
  Mental Retardation and Developmental Disabilities Research Reviews
  \textbf{11}(1),  61--67 (2005)

\bibitem{farneback2003two}
Farneb{\"a}ck, G.: Two-frame motion estimation based on polynomial expansion.
  In: Proceedings of the Scandinavian Conference on Image Analysis. pp.
  363--370. Springer (2003)

\bibitem{feichtenhofer2016convolutional}
Feichtenhofer, C., Pinz, A., Zisserman, A.: Convolutional two-stream network
  fusion for video action recognition. In: Proceedings of the IEEE Conference
  on Computer Vision and Pattern Recognition (CVPR). pp. 1933--1941 (2016)

\bibitem{ferrari1990qualitative}
Ferrari, F., Cioni, G., Prechtl, H.: Qualitative changes of general movements
  in preterm infants with brain lesions. Early Human Development
  \textbf{23}(3),  193--231 (1990)

\bibitem{he2016deep}
He, K., Zhang, X., Ren, S., Sun, J.: Deep residual learning for image
  recognition. In: Proceedings of the IEEE Conference on Computer Vision and
  Pattern Recognition (CVPR). pp. 770--778 (2016)

\bibitem{hesse2018learning}
Hesse, N., Pujades, S., Romero, J., Black, M.J., Bodensteiner, C., Arens, M.,
  Hofmann, U.G., Tacke, U., Hadders-Algra, M., Weinberger, R., et~al.: Learning
  an infant body model from {RGB-D} data for accurate full body motion
  analysis. In: Proceedings of the International Conference on Medical Image
  Computing and Computer-Assisted Intervention (MICCAI). pp. 792--800 (2018)

\bibitem{loshchilov2017decoupled}
Loshchilov, I., Hutter, F.: Decoupled weight decay regularization. In:
  Proceedings of the International Conference on Learning Representations
  (ICLR) (2017)

\bibitem{maggi2017new}
Maggi, E., Impagliazzo, M., Minnella, A., Zanardi, N., Izzo, M., Campone, F.,
  Blanchi, I., Tacchino, C., Shimatani, K., Shima, K., et~al.: A new method for
  early detection of infants at risk of long-term neuromotor disabilities. Gait
  \& Posture  \textbf{57},  23--24 (2017)

\bibitem{mccay2020abnormal}
McCay, K.D., Ho, E.S., Shum, H.P., Fehringer, G., Marcroft, C., Embleton, N.D.:
  Abnormal infant movements classification with deep learning on pose-based
  features. IEEE Access  \textbf{8},  51582--51592 (2020)

\bibitem{nguyen2021spatio}
Nguyen-Thai, B., Le, V., Morgan, C., Badawi, N., Tran, T., Venkatesh, S.: A
  spatio-temporal attention-based model for infant movement assessment from
  videos. IEEE Journal of Biomedical and Health Informatics  \textbf{25}(10),
  3911--3920 (2021)

\bibitem{prechtl1990qualitative}
Prechtl, H.F.: Qualitative changes of spontaneous movements in fetus and
  preterm infant are a marker of neurological dysfunction. Early Human
  Development  \textbf{23}(3),  151--158 (1990)

\bibitem{prechtl1997state}
Prechtl, H.F.: State of the art of a new functional assessment of the young
  nervous system. an early predictor of cerebral palsy. Early Human Development
   \textbf{50}(1),  1--11 (1997)

\bibitem{qin2020u2}
Qin, X., Zhang, Z., Huang, C., Dehghan, M., Zaiane, O.R., Jagersand, M.:
  {U2-Net}: {Going} deeper with nested {U}-structure for salient object
  detection. Pattern Recognition  \textbf{106},  107404 (2020)

\bibitem{schmidt2019general}
Schmidt, W., Regan, M., Fahey, M., Paplinski, A.: General movement assessment
  by machine learning: Why is it so difficult. Journal of Medical Artificial
  Intelligence  \textbf{2} (2019)

\bibitem{simonyan2014two}
Simonyan, K., Zisserman, A.: Two-stream convolutional networks for action
  recognition in videos. In: Proceedings of the 27th International Conference
  on Neural Information Processing Systems (NIPS). pp. 568--576 (2014)

\bibitem{tacchino2021spontaneous}
Tacchino, C., Impagliazzo, M., Maggi, E., Bertamino, M., Blanchi, I., Campone,
  F., Durand, P., Fato, M., Giannoni, P., Iandolo, R., et~al.: Spontaneous
  movements in the newborns: A tool of quantitative video analysis of preterm
  babies. Computer Methods and Programs in Biomedicine  \textbf{199},  105838
  (2021)

\bibitem{tran2015learning}
Tran, D., Bourdev, L., Fergus, R., Torresani, L., Paluri, M.: Learning
  spatiotemporal features with {3D} convolutional networks. In: Proceedings of
  the IEEE International Conference on Computer Vision (ICCV). pp. 4489--4497
  (2015)

\bibitem{tsuji2020markerless}
Tsuji, T., Nakashima, S., Hayashi, H., Soh, Z., Furui, A., Shibanoki, T.,
  Shima, K., Shimatani, K.: Markerless measurement and evaluation of general
  movements in infants. Scientific Reports  \textbf{10}(1), ~1422 (2020)

\end{thebibliography}
\bibliographystyle{splncs04}

\end{document}